\documentclass[[a4paper]{article}

\usepackage{graphicx}%
\usepackage{multirow}%
\usepackage{amsmath,amssymb,amsfonts}%
\usepackage{amsthm}%
\usepackage{mathrsfs}%
\usepackage[title]{appendix}%
\usepackage{xcolor}%
\usepackage{textcomp}%
\usepackage{manyfoot}%
\usepackage{algorithm}%
\usepackage{listings}%
\usepackage{natbib}

\usepackage{caption}

\newcommand{\ns}{n}
\newcommand{\nvs}{p}
\newcommand{\nl}{m}

\newcommand{\V}{U}
\newcommand{\W}{V}

\newcommand{\ww}{v}
\newcommand{\vv}{u}
\newcommand{\ve}{\nu}

\newcommand{\polar}{\mathrm{polar}}
\newcommand{\spann}{\mathrm{span}}
\newcommand{\rank}{\mathrm{rank}}
\newcommand{\diag}{\mathrm{diag}}
\newcommand{\tr}{\mathrm{tr}}
\newcommand{\pr}{\mathrm{P}}
\newcommand{\p}{^{\prime}}

\newcommand{\pev}{\mathrm{pev}}
\newcommand{\var}{\mathrm{expvar}}
\newcommand{\varpca}{\sigma^{2}_{1}+\dots+\sigma^{2}_{\nl}}
\newcommand{\varsubspace}{\mathrm{\var}_{subsp}\,}
\newcommand{\varproj}{\mathrm{\var}_{proj}}
\newcommand{\varprojqr}{\mathrm{\var}_{proj}^{QR}}
\newcommand{\varprojup}{\mathrm{\var}_{proj}^{UP}}
\newcommand{\varprojopt}{\mathrm{\var}_{proj}^{opt}}
\newcommand{\varprojoptmu}{\mathrm{\var}_{proj,\mu}^{opt}}
\newcommand{\varnorm}{\mathrm{\var}_{norm}}
\newcommand{\varnormqr}{\mathrm{\var}^{QR}_{norm}}
\newcommand{\varnormup}{\mathrm{\var}^{UP}_{norm}}

\newcommand{\R}{I\!\!R}
\newcommand{\E}{ \mathcal{E}}

\newcommand{\EX}{\mathcal{E}^{X}}
\newcommand{\PX}{\mathcal{P}^{X}}

\newcommand{\cqfd}{\mbox{}\hfill\rule{.8em}{1.8ex}}
\newcommand{\half}{\frac{1}{2}}
\newcommand{\egaldef}{\stackrel{\mathrm{def}}{=}}
\newcommand{\egalperm}{\, \stackrel{\mathrm{perm}}{=}  \,}
\newcommand{\egalquest}{\stackrel{\mathrm{?}}{=}}

\newtheorem{Theorem}{Theorem}[section]
\newtheorem{Lemma}[Theorem]{Lemma}
\newtheorem{Proposition}[Theorem]{Proposition}

\author{
  Marie Chavent 
  \thanks{Universit\'e de Bordeaux, CNRS, INRIA, Bordeaux INP, IMB, UMR 5251, Talence, France
  \newline \hspace*{1,5em} e-mail~:   {\tt marie.chavent@u-bordeaux.fr} (corresponding author)}
 \hspace{0,2em}
  \and Guy Chavent
  \thanks{Retired, collaborator to the SERENA  Project Team, INRIA-Paris  2 rue Simone Iff, 75589 Paris, France}
}

\title{From explained variance of correlated components to PCA without orthogonality constraints}

\date{}

\begin{document}

\begin{titlepage}

\maketitle
\thispagestyle{empty}
\setcounter{page}{0}

\begin{abstract}
Block Principal Component Analysis (Block PCA) of a data matrix $A$, where loadings $Z$ are determined by maximization of $\|AZ\|^{2}$ over unit norm  orthogonal loadings,
is difficult to use for the design of sparse PCA by $\ell^{1}$ regularization, due to the difficulty of taking care of both the orthogonality constraint on loadings and the non differentiable $\ell^{1}$ penalty. Our objective in this paper is to relax the orthogonality constraint on loadings by introducing new objective functions $\var(Y)$ which measure the part of the variance of the data matrix $A$ explained by correlated components $Y=AZ$. 
So we propose first a comprehensive study of  mathematical and numerical properties of $\var(Y)$ for two existing definitions \cite{zou2006sparse},  \cite{SH2008} and four new definitions. Then we show that only two of these explained variance are fit to use as objective function in block PCA formulations for $A$ rid of orthogonality constraints. 
\end{abstract}

\noindent
\textbf{Keywords:}  PCA, sparsity, dimension reduction, explained variance, orthogonality constraints, block optimization.

\end{titlepage}

\section{Introduction}\label{intro}

Many linear data analysis methods construct new variables that ``best'' summarize the columns of a $n \times p$ data matrix $A$ where $n$ observations are described by $p$  numerical variables. The  $m <p$ new variables are the columns of $Y=AZ$  where $Z$ is the $p \times m$ loading matrix. 
%defined according to the optimization problem under consideration. 
These new variables are for instance principal components in PCA (Principal Component Analysis), canonical components in CA (Canonical Analysis) or PLS components in PLS (Partial Least Squares) regression. When the components are orthogonal (which is the case for PCA), it is usual to assess the quality of the components by measuring the part of the variance of $A$ explained  by $Y=AZ$ with~:
$$ \var(AZ) = \|AZ\|^{2}_{F},$$
where the subscript $F$ denotes the Frobenius norm.

But this definition fails when the components are correlated. Correlated components appear when sparsity is introduced into the loading matrix $Z$ to select the important original variables.  The loading vectors %, and therefore the components,
and the components are no longer necessarily orthogonal. Two definitions have already been proposed to measure the explained variance of correlated components~: the {\em adjusted variance}  of  \citet{zou2006sparse} and the {\em total variance} of \cite{SH2008}. Because there is no single definition, we introduce a set of three conditions to be satisfied by any explained variance definition and we propose a comprehensive study of  mathematical and numerical properties of  these two existing definitions together with four new definitions. 

We prove first that the  \emph{total variance} 
of~\cite{SH2008} called \emph{subspace explained variance} hereafter, satisfies only two of three conditions. This lead us 
to propose other definitions by \emph{projection} or by \emph{normalization} on a set $X=[x_1,...,x_m]$ of orthonormal vectors which ``points in the same direction'' as the components $Y=[y_1,...,y_m]$. Such orthonormal vectors $X$ can be associated to $Y$ by three different rules: QR or UP (polar) decomposition of $Y$, or maximization of the explained variance. This leads to five definitions ( two normalized  and three projected explained variance)~:

\begin{itemize}
 \item[-] the QR and the UP normalized explained variances (QRnormVar and UPnormVar),
 \item[-] the QR~projected explained variance (QRprojVar) which is the adjusted variance of~\cite{zou2006sparse},
 \item[-] the UP~projected explained variance (UPprojVar),
 \item[-] the optimal projected explained variance (optprojVar).

\end{itemize}
We prove that the five above definitions satisfy also the two first compatibility conditions, but that only the three projected variances, satisfy the third one, and hence are proper explained variances.

 Then we investigate both theoretically and numerically the size of the six explained variances, the existence of order between them, and their ability to rank in the same order - or not - different sets of components $Y$, which is for what they have been introduced at the origin.

Finally, we study the ability of the three proper definitions of explained variance to 
provide a substitute to the classical Block PCA formulation~:
\begin{equation}
\label{583}
\text{maximizing  $\|AZ\|^{2}_{F}$ under the orthonormality constraint
$Z^{T}Z = I_{\nl}$,}
\end{equation}
by replacing \eqref{583} by a   \emph{Explained Variance block PCA formulation}~:
\begin{equation}
\label{cp1-10}
\displaystyle \max_{ \|z_{j}\|=1 , j=1\dots \nl }
   \var(AZ)  \ ,
\end{equation}
which is rid of orthogonality constraints on loadings $Z=[z_{1}\dots z_{\nl}]$, and hence particularly suited as starting point for the design of sparse PCA algorithms.. 

 As a conclusion of this study, we single out  the {\em optimal projected explained variance} of $Y=AZ$~:
\begin{equation}
\label{cp1-15}
\var(Y) =  \max_{X^{T}X=I_{\nl}} \sum_{j=1\dots\nl} \langle y_{j}\,,x_{j}\rangle^{2} \ ,
\end{equation}
which possesses, after introduction of weights, the desired properties in the sense that it is easy to compute and differentiate, and admits as unique maximizer the SVD solution of PCA $Z^{*} = [v_{1} \dots v_{\nl}]$ made of the $\nl$ first right singular vectors of $A$. This will be the starting point for the design of an efficient group-sparse block PCA algorithm  in the companion paper \citep{chavent2023-cp-two}.

%With the objective of complementing the  {\em block PCA formulation}~:
%\begin{equation}
%\label{cp1-5} \nonumber
% \max_{Z^{T}Z = I_{\nl}} \|AZ\|^{2}_{F} \ \ 
%\end{equation}

The paper is organized as follows: In Section~\ref{how to define the explained variance} we motivate the need for an explained variance definition, and establish necessary conditions to be satisfied by any definition of the variance explained by a set of non-necessarily orthogonal components. Section \ref{section: subspace variance} establishes the properties of the subspace explained variance (total variance) of \cite{SH2008}. Section \ref{adjusted optimal and orthogonal variances} is devoted to the definition of three projected (including the adjusted variance of \cite{zou2006sparse} and the optimal variance \eqref{cp1-15}) and two normalized explained variances, together with the study of their mathematical properties. Sections \ref{comparison of explained variances} and \ref{ranking properties of explained variances} present numerical comparisons of the size of the variance explained by the six definitions, and of their ability to rank in the same order (or not) non necessarily orthogonal components. Finally, Section \ref{explained variance block pca formulations} compares the ability of the three projected variance definitions 
%from the points of view of magnitude, computational and differentiation ease, 
to discriminate the singular value solution of PCA as their unique maximizer, which makes them fit to use as Block PCA objective function.

It should be noted that we have implemented the six explained variance definitions in R \citep{rsoftware} in the package \textbf{sparsePCA} available at \texttt{https://github.com/chavent/sparsePCA}.

%%%-----------

\section{Defining variance explained by components when loadings are non orthogonal }
\label{how to define the explained variance}

We set us from now on in the context of Principal Component Analysis (PCA), where one seeks a small number $\nl$ of uncorrelated components $y_{j}=Az_{j}, j=1\dots \nl$ by combining the $\nvs$ columns of a data matrix $A$, each containing $\ns$ samples of a centered variable, with $\nl$ unit norm \emph{loading} vectors 
$z_{j},j=1\dots \nl$, in such a way that the components $y_{j}$ retain the largest possible part of the total variance of $A$. We denote by 
\begin{equation}
\label{101}
   \var(Y) = \text{the  part of the variance of $A$ explained by $Y=AZ$} \ 
\end{equation}
and by
\begin{equation}
\label{101a }\nonumber
   \|A\|^{2}_{F} = \sum_{i,j = 1\dots \nl} a_{i,j}^{2} \text{ the (total) variance of $A$, where $\| \bullet \|_{F}$ is the Frobenius norm of $A$} \ .
\end{equation}
At this point, $\var(Y)$ is a still loosely defined quantity outside of the solution of PCA given by~:
 \begin{equation}
\label{110}
   \quad Z= \W_{\nl} , \quad  Y= \V_{\nl}\, \Sigma_{\nl}=A\W_{\nl}  \ ,
\end{equation}
where  $\V_{\nl}$ and $\W_{\nl}$ contain the $\nl$ first left and right singular vectors and $\Sigma_{\nl}$ is the diagonal matrix of the $\nl$ first singular values of the
 singular value decomposition (SVD) of $A$~:
\begin{equation}
\label{106}\nonumber
\begin{array}{l}
    A=\V\Sigma \W^{T}\quad \mbox{ with } \quad \V^{T}\V=I_{r}\quad,\quad 
    \W^{T}\W=I_{r} \ , \\
 \Sigma=\diag(\sigma_{1}, \dots,\sigma_{r}) = \mbox{$r \times r$ matrix with } \sigma_{1}\geq \sigma_{2}\geq \dots \geq \sigma_{r} >0 \ ,
 \end{array}
\end{equation}
where $r$ is the rank of $A$,
and the columns $\vv_{1} \dots \vv_{r}$ of $\V$ and  $\ww_{1} \dots \ww_{r}$ of $\W$ are the left and right singular vectors of $A$.

The principal components $y_{j}= Az_j=\sigma_{j} u_{j}$ are orthogonal, and hence uncorrelated, so the sum of their variance  $\|y_{j}\|^{2}$ represents the part of the total variance $\|A\|_{F}^{2}$ of $A$ explained by these $\nl$ principal components. So we see that  the variance explained by the principal components $Y$ is~:
\begin{equation}
\label{115}
  \var(Y)  = \|Y\|_{F}^{2} = \sum_{j=1 \dots \nl} \|y_{j}\|^{2} 
  	 =\sum_{j=1 \dots \nl} \sigma_{j}^{2}                                  
   	\ \leq \ \sum_{j=1 \dots r} \sigma_{j}^{2}
   	=  \|A\|_{F}^{2} \ .
\end{equation}
However, when sparsity constraints are introduced into loading vectors  like in sparse PCA for instance, non orthogonal loadings and components are generated and one has to face the problem of defining $\var(Y)$ for possibly non orthogonal components. Alas, formula \eqref{115} for $\var(Y)$ is strictly limited to the case of orthogonal components and loadings, as we see now.

Consider first the case of non orthogonal components~:  take for example for $Z$  an orthonormal basis of $\spann\{\W_{\nl}\}$  but different from $\W_{\nl}$. Then components $Y=AZ$ are not anymore orthogonal, and hence correlated, so the sum of their variances (the total variance of $Y$) is too optimistic, and one expects that
\begin{equation}
\label{105-11}
   \var(Y) < \|Y\|^{2}_{F} \ ,
\end{equation}
which shows that $\|Y\|^{2}_{F}$ is not a satisfying definition of $\var(Y)$ in this case.

Then take for example a data matrix $A$ with three singular values $3,2,1$ and hence a total variance of 14. Then chose for $Z$ two linearly independant \emph{unit vectors} close to the first right singular vector $\ww_{1}$. Then~:
\begin{equation}
\label{528}\nonumber
 \|AZ\|_{F}^{2}= \|Y\|_{F}^{2}= \|y_{1}\|^{2}+\|y_{2}\|^{2}= \underbrace{\|Az_{1}\|^{2}}_{\simeq \sigma_{1}^{2}=9} +  \underbrace{\|Az_{2}\|^{2}}_{\simeq \sigma_{1}^{2}=9} \simeq 18 > \underbrace{9+4+1}_{\sigma_{1}^{2}+\sigma_{2}^{2}+\sigma_{3}^{2}}=14 =\|A\|_{F}^{2} \ ,
\end{equation}
which violates property $\var(Y) \leq \|A\|^{2}_{F}$ implied by \eqref{101},  so once again $\|Y\|^{2}_{F}$ is not suited as a definition of $\var(Y)$.

A last motivation for the search of definitions of $\var(Y)$~: 
%is to use them as objective function for the design of new explained variance block  PCA formulations~: 
rather than solving the classical Block PCA formulation \eqref{583},
%\begin{equation}
%\label{583}
%\text{maximizing  $\|AZ\|^{2}_{F}$ under the orthonormality constraint
%$Z^{T}Z = I_{\nl}$,}
%\end{equation}
why not solve the explained variance Block PCA formulation \eqref{cp1-10} by maximizing $\var(Y)$ under the sole unit norm constraint on the loadings $Z$~? 
%This will be developped in Section \ref{explained variance block pca formulations} below. 
Sparse block PCA formulations based on such explained variance objective function eliminate the difficulty caused by the orthogonality constraints on the loadings, and, by construction, rule directly the balance between sparsity and explained variance.  

% will make possible the design of sparse explained variance block PCA formulation  \cite{chavent2023-cp-two} which will rule directly the balance between sparsity and explained variance.  

\vspace{2ex}

So we address first in this paper the problem of defining the part of the variance of $A$ explained by components $Y=AZ$, under the sole condition that~:
\begin{equation}
\label{500}
  \|z_{j}\| = 1,j=1\dots \nl    \quad  , \quad   \rank(Z) =\rank(Y)=\nl  \leq r \ .%\egaldef \rank A   \,
\end{equation}
In absence of a sound definition for the explained variance of correlated components, we  define a set of hopefully reasonable necessary conditions to be satisfied by any such definition~:
\begin{itemize}
  \item \textbf{Condition 1}~: when $Y,Z$ happen to be the SVD solution of PCA given by \eqref{110}, $\var(Y)$ has to provide the \emph{exact value} given by \eqref{115}~:
\begin{equation}
\label{510}
  \var(Y) =  \sum_{j=1 \dots \nl} \sigma_{j}^{2}                                  
   \ \leq \ \sum_{j=1 \dots r} \sigma_{j}^{2}  =  \|A\|_{F}^{2} \quad \text{ for } m=1 \dots r \ .
\end{equation}
  \item \textbf{Condition 2}~: when $Y,Z$ satisfy only \eqref{500}, the components are not orthogonal anymore, and one expects that, because of the correlation between the components, the part of the variance of $A$ explained by $Y$ is smaller than that of the PCA solution~:
 \begin{equation}
\label{510a}
\var(Y) \leq \sum_{j=1 \dots \nl} \sigma_{j}^{2}    \ .
\end{equation}
  \item \textbf{Condition 3}~: $\var(Y)$ has to take into account the possible correlation of the components as expected in \eqref{105-11}~:
\begin{equation}
\label{504}\nonumber
  \var(Y) \leq \sum_{j=1\dots \nl} \|y_{j}\|^{2} = \|Y\|_{F}^{2} \ , \ 
\end{equation}
with equality only if the components are orthogonal.
\end{itemize}
 Any explained variance definition $\var(AZ)$
 which satisfies these conditions achieves its PCA maximum value 
 $\sum_{j=1 \dots r} \sigma_{j}^{2}$ (Condition 2) for $Z=\W_{\nl}$ (Condition 1), and hence  provides a block formulation for PCA  without orthogonality constraints on loadings, which can be used to derive sparse PCA algorithms. This point os view will be developped in Section \ref{explained variance block pca formulations} below.

We propose now definitions for the explained variance  of  the components $Y=AZ$ associated to any matrix $Z$ of $m\leq r$ linearly independant - but not necessarily orthogonal - unit norm loading vectors $z_j$, $j=1,...,m$.
These definitions will include those introduces by \cite{SH2008}  (see section \ref{section: subspace variance} below) and \cite{zou2006sparse} (see section \ref{section: projected explained variance}).

%%%-----------

 \section{Subspace explained variance}
 \label{section: subspace variance}

 We start here from a reformulation of the explained variance \eqref{115}
 of the principal components $Y=AZ$, based on the subspace spanned by $Z=\W_{\nl}$. 
 Let $ \pr_{\W_{\nl}}$ denotes the orthogonal projection on this subspace. 
 Then  $\pr_{Z} = \pr_{\W_{\nl}} = \W_{\nl} \W_{\nl}^{T}$, so that~:
\begin{equation}
\label{515}
    \var(Y) =\|Y\|_{F}^{2} =  \|A \,\pr_{Z}\|_{F}^{2} = 
    \|A \,\pr_{\W_{\nl}}\|_{F}^{2} =\sum_{j=1 \dots \nl} \sigma_{j}^{2}  \ .
\end{equation}
When $Z$ satisfies (\ref{500}) only, we proceed  by analogy with  (\ref{515}), and define the \emph{subspace explained variance}  of $Y=AZ$ by~:
\begin{equation}
\label{540}
    \varsubspace(Y) \egaldef  \|A\pr_{Z}\|_{F}^{2}
    =  \tr\big\{Y^{T}Y(Z^{T}Z)^{-1})\big\} \ ,
\end{equation}
where $\pr_{Z}= Z (Z^{T}Z)^{-1}Z^{T}$ denotes the projection matrix on the subspace spanned by $Z$.
This shows that subspace explained variance is the Rayleigh quotient associated to $A,Z$.
Note that with this definition, where the explained variance depends only of the subspace spanned by $[z_{1} \dots z_{\nl}]$, the normalization of the loadings $z_{j}$ is not necessary. Of course, we will still continue to represent loadings by unit norm vectors - but this is here only a convenience.

The subspace explained variance coincides with the \emph{total explained variance} introduced by \citet[section 2.3 p. 1021]{SH2008}, which they proved was increasing with the number of loadings, and bounded by the variance $\|A\|^{2}_{F}$ of the data. The next lemma gives a complete picture of its properties~:

\medskip

\begin{Lemma} \emph{(Subspace  explained Variance)}
\label{lem 1}
Let $Z$ satisfy (\ref{500}). Then the subspace explained variance defined by \eqref{540} satisfies~:
\begin{equation}
\label{546-2}
  \varsubspace(Y) = 
  %\!\! \sum_{j=1 \dots \nl} \sigma_{j}^{2} = \emph{Max}
   \!\! \varpca = \emph{Max}
   \quad \quad \Leftrightarrow \quad \quad  \spann\{Z\} = \spann\{\W_{\nl}\}   \ .
\end{equation}
and hence satisfies conditions 1 and 2. However, it does not satisfy condition 3~:
\begin{itemize}
  \item[-] when the components $Y=AZ$ happen to be orthogonal~:
 \begin{equation}
\label{546-1}
 \varsubspace(Y)    \geq  \|Y\|_{F}^{2}  \  
 \emph{ with equality iff } Z  \egalperm \W_{\nl} \ ,
\end{equation}
where ``$\egalperm$'' denotes the equality of matrices up to a column permutation and multiplicatipn by $\pm1$,
\smallskip
  \item[-]  when the loadings $Z$ happen to be orthogonal~:
\begin{equation}
\label{546-3}
\varsubspace(Y) = \|Y\|_{F}^{2} \quad \text{with non necessarily orthogonal components} \ .
\end{equation}
\end{itemize}
\end{Lemma}

The proof is in Section \ref{proof of lem 2} of the Appendix. 
This lemma shows that $\varsubspace(Y)$  verifies only conditions 1 and 2, and overestimates the explained variance in two cases~:
\begin{itemize}
  \item[-]  when components $Y$ are orthogonal without pointing in the direction of the left singular vectors, inequality \eqref{546-1} implies $\varsubspace(Y)> \|Y\|_{F}^{2}$, which contradicts condition 3,
  \item[-]  when loadings $Z$ are orthogonal without pointing in the direction of the right singular vectors, then the components are correlated and \eqref{546-3} contradicts condition 3, which requires in that case that  $\varsubspace(Y) < \|Y\|_{F}^{2}$.
\end{itemize}

We explore in the next section other directions in the hope of being able to comply with all conditions 1, 2 and 3.

%-----------------------------

\section{Projected  and normalized explained variances}
\label{adjusted optimal and orthogonal variances}

We start now from definition (\ref{115}) of the explained variance in the case of PCA. A natural generalization
 would be~:
\begin{equation}
\label{520}\nonumber
  \var(Y) \egalquest \sum_{j=1 \dots \nl} \|y_{j}\|^{2} =\|Y\|_{F}^{2} = \|AZ\|_{F}^{2}  \ ,
\end{equation}
which, as we have seen in the Section \ref{how to define the explained variance}, is not anymore an an acceptable definition when components are correlated.

\begin{figure}[h]
\begin{center}
\centerline{\resizebox{28em}{!}{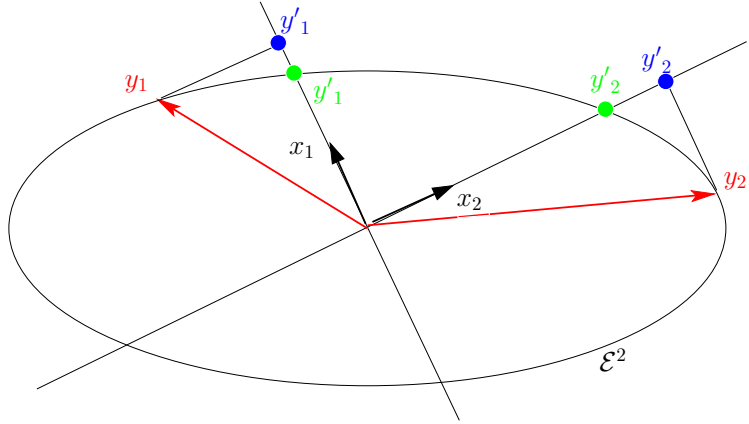}}
\caption{Illustration of \textcolor{blue}{ projected} and \textcolor{green}{normalized} explained variances~: The ellipse  $\E^{2}$ represents the image by $A$ of all unit norm loadings.
Let the $X=[x_{1},x_{2}]$ be the orthonormal vectors associated to the correlated components $Y=[y_{1},y_{2}]$. Then the extremities of the adjusted components $Y\p = [y\p_{1},y\p_{2}]$ obtained by \textcolor{blue}{projection} are the two \textcolor{blue}{blue dots} on the $x_{1},x_{2}$ axes, and those obtained by \textcolor{green}{normalization} are
the two  \textcolor{green}{green dots} located at the intersection of the axes with  $\E^{2}$.}
\label{fig 1a}
\end{center}
\end{figure}

However, this definition continues to make perfect sense for the variance explained by components as long as they are orthogonal, without pointing necessarily in the direction of left singular vectors.
Hence a natural way to eliminate the redundancy caused by the orthogonality default of the components $Y$  \emph{and} to satisfy Condition 3 is to~:
\begin{enumerate}
  \item \textbf{Step 1}: choose a rule to associate to the components $Y$ an \emph{orthonormal basis} $X$ of $\spann\{Y\}$ that, loosely speaking, ``points in the direction of the components~$Y$'',
and, when the components $Y$ happen to be 
orthogonal, points in the directions of $Y$  itself.
So the rule for the choice of the basis $X$ associated to $Y$ has to satisfy~:
\begin{equation}
\label{530}
\begin{array}{l}
     X^{T}X = I_{\nl} \quad, \quad \spann\{X\} = \spann\{Y\} \ ,      \\
      <y_{j},x_{j}> \ \geq 0 \ \ \forall j=1\dots \nl \ ,  \\
    \langle y_{j},y_{k}\rangle = 0  \  \forall  j \neq k \quad  \Longrightarrow
     \quad x_{j}=y_{j}/\|y_{j}\|  \  \forall j=1 \dots \nl \ .
\end{array}
\end{equation}
%We denote by $M$ the matrix of the coordinates of $Y$ in the chosen basis~:\begin{equation}
%\label{530-2}
%   M = X^{T}Y \quad \Longleftrightarrow  \quad Y = XM \ ,
%\end{equation}
%which is regular as the components $Y$ are supposed linearly independent.
Examples of such rules are~: 
\begin{equation}
\label{531}
\begin{array}{llll}
  \text{QR decomposition~:} &Y \leadsto X=Q & \text{ solution of } &
   Y = Q\, R \  ,\   Q^{T}Q = I_{\nl}  \quad \\
  & \text{where  $R$ is an upper triangular matrix} \ ,
  \hspace{-15em}   \\
  \text{Polar decomposition~:} & Y \leadsto X=U & \text{ solution of } &  
   Y = U\, P \  ,\   U^{T}U = I_{\nl} \\
   & \text{where }  P^{T} \!= P \in \R^{\nl \times \nl} \  , \  P \geq 0  \ .\hspace{-15em} 
\end{array}
\end{equation}
\item \textbf{Step 2}: associate to $Y$ \emph{orthogonal adjusted components} $Y\p$ along the $X$ axes, and define the variance explained by the components $Y$ by~:
\begin{equation}
\label{533-1}\nonumber
  \var(Y) \egaldef \|Y\p\|_{F}^{2} \ .
\end{equation}  
\end{enumerate}
Two ways for obtaining the adjusted components $Y\p$ are considered hereafter: by projection or by normalization as illustrated in Figure \ref{fig 1a}.

%-----------------------------

\subsection{Normalized explained variances}

We consider first in this section the case where the adjusted components $Y\p$ in step 2 are obtained by ``normalization'' in the directions of the chosen orthonormal basis  $X$ of $\spann\{Y\}$.
More precisely, the idea is to choose  the abscissa of $y\p_{j}$ on the $x_{j}$ axis by requiring that $y\p_{j}$ is the image by $A$ of some unit norm \emph{adjusted  loading} $z\p_{j}$. 
This is illustrated in Figure \ref{fig 1a}~: one associates to components $Y=[y_{1},y_{2}]$ the adjusted components
$Y\p = [y\p_{1},y\p_{2}]$ whose extremities are the points on the $x_{1},x_{2}$ axes
located on the ellipse image by $A$ of the unit ball of the loading space. In order to determine the adjusted loadings $z\p_{j}$, one computes first (non necessarily unit norm) loadings $T=[t_{1},\dots,t_{\nl}]$ by performing on the loadings $z_{j}$ the same linear combinations~$M$~:
 \begin{equation}
\label{534-11-1}
Y=XM \ ,
\end{equation}
that transformed $Y$ into $X$, which leads to define $T$ by~:
\begin{equation}
\label{534-13}
    Z = T M \ .% \quad \quad \text{(compare to~: $Y=XM$)} \ .
\end{equation}
Multiplying then by $A$ left gives $Y=AZ=ATM$ and, comparison with \eqref{534-11-1} shows that $X=AT$ and so $x_{j}=At_{j},j=1\dots \nl$.
The adjusted components ${y\p}_{j}$ are then defined~by~:
 \begin{equation}
\label{532-1}%\nonumber
   {y\p}_{j} = A {z\p}_{j} \ \   \mbox{with}\ \   z\p_{j}=t_{j}/\|t_{j}\|
   \ \  \text{ so that }
   \ \  \|z\p_{j}\| = 1 \ ,\  y\p_{j}= x_{j}/\|t_{j}\|
    \ , \  j=1\dots\nl \ ,
\end{equation}
and the \emph{ normalized explained variance} of $Y$ estimated with $X$ is  defined by~:
\begin{equation}
\label{539}
   \varnorm^{X}(Y) = \|Y\p\|^{2}_{F} =  \sum_{j=1\dots\nl} 1/\|{t}_{j}\|^{2} \ ,
\end{equation}
where $T=[t_{1},\dots,t_{\nl}]$ is given by \eqref{534-13} and \eqref{534-11-1}. Before specifying the rule of step 1 which  associates  an orthonormal basis $X$  to the components $Y$, we give some properties of $\varnorm^{X}(Y)$  which hold independently of the chosen basis $X$.

\medskip

 \begin{Lemma} \emph{(Normalized explained variance)}
\label{lem 4}
For any unit norm loadings $Z$ and any basis $X$ chosen according to the rule \eqref{530}, the normalized explained variance  of $Y=AZ$ defined by\eqref{539}  satisfies conditions 1 and 2 and~:
%Let $Y=AZ$ be components  associated to $\nl$ unit  norm loading vectors $Z=(z_{1}, \dots ,z_{\nl})$. The for any choice of  $X$ which satisfies the rule \eqref{530}, the normalized explained variance $\varnorm^{X}(Y)$ defined by \eqref{539}  satisfies conditions 1, 2 and 3 and~:
\begin{equation}
\label{534-3}
     \varnorm ^{X}(Y)  \leq   \varsubspace(Y) \leq \varpca \ .
 \end{equation}
 \end{Lemma}

\noindent  Lemma \ref{lem 4} follows from Lemma \ref{lem 1} applied to the orthogonal components $Y\p$~:
\begin{equation}
\label{534-43}
 \varnorm^{X}(Y) \egaldef \|Y\p\|^{2}_{F} \leq \varsubspace(Y\p) 
 %= \varsubspace(Y)
  \leq \varpca \ ,
\end{equation} 
which proves \eqref{534-3}. We can now specify the rules for the choice of $X$ to define two normalized explained variance satisfying conditions 1 and 2.

\begin{figure}[h]
\begin{center}
\centerline{\resizebox{28em}{!}{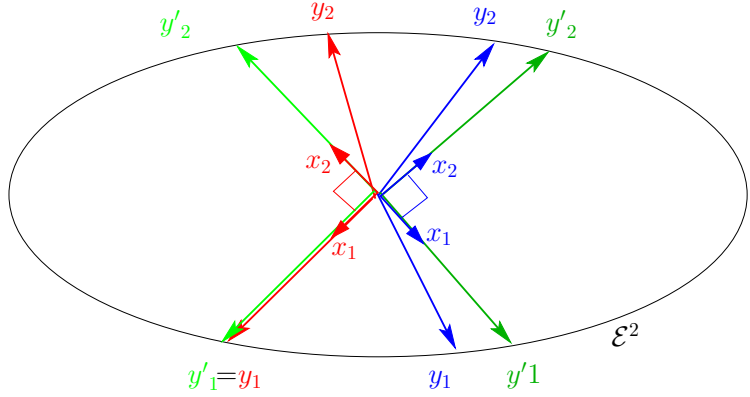}}
\caption{Counterexamples for property 3. 
\textbf{Left~:} Let \textcolor{red}{$X$} denote the basis associated to components 
 \textcolor{red}{$Y$} 
 by the QR-decomposition, and  \textcolor{green}{$Y\p$} be the
corresponding normalized adjusted components. One sees that 
$\textcolor{green}{\varnormqr(Y)
= \|Y\p\|^{2}} \geq  \textcolor{red}{\|Y\|^{2}}$,
which violates property~3.
\textbf{Right~:} Let \textcolor{blue}{$X$} denote the basis associated to components \textcolor{blue}{$Y$} 
 by the polar decomposition, and   \textcolor{olive}{$Y\p$} be the
corresponding normalized adjusted components. One sees that 
$\textcolor{olive}{\varnormqr(Y)
= \|Y\p\|^{2}} \geq  \textcolor{blue}{\|Y\|^{2}}$,
which violates property~3.
}
\label{fig 1c}
\end{center}
\end{figure}

\paragraph{QR normalized variance.} 
Let  $X$  be defined by the QR-decomposition $Y = XR$  of the components $Y = AZ$ as recalled in \eqref{531} . Then \eqref{539} leads to another definition of variance~:
\begin{equation}
\label{539-0}%\nonumber
 \hspace{-0,5em} \varnormqr(Y) = %\tr\{ \diag^{-1}(T^{T}T)  \} 
  \!\!\!\! \sum_{j=1\dots\nl} \!\!1/\|{t}_{j}\|^{2} 
 \   \ , %\mbox{ with } \   Y = XR ,  \text{ $R$ upper triangular, and } 
   \quad Z=TR\ ,\ \text{$R$ upper triangular .} 
  \end{equation}
  The normalized variance $\varnormqr(AZ)$ does not satisfy condition 3, as the counter example of Figure \ref{fig 1c}, left, shows.

\paragraph{UP normalized variance.} Let $X$ be defined by the UP-decomposition (polar decomposition) $Y=XP$  of the components $Y=AZ$ as recalled in \eqref{531}. Then 
 (\ref{539}) defines another variance~:
\begin{equation}
\label{539-1}%\nonumber
  \varnormup(Y)\! =
   %\tr\{ \diag^{-1}(T^{T}T)  \}  
   \!\!\!\! \sum_{j=1\dots\nl} \!\!1/\|{t}_{j}\|^{2} \ \ ,
    \quad Z=TP \  \  , \ \ P=(Y^{T}Y)^{1/2} \ .
    %\ \mbox{ with } Y=XP , \ P=(Y^{T}Y)^{1/2} ,  \text{ and }  Z=TP \ ,
  % T=Z(Y^{T}Y)^{-1/2} \ ,
  %\hspace{-2em}
  \end{equation}
 The normalized variance $\varnormup(AZ)$ does not either satisfy condition 3, as the counter example of Figure \ref{fig 1c}, right, shows.
 
\smallskip
 
So we explore in the next section another road in order to comply with all conditions 1,2 and 3.

%%%------------------------

\subsection{Projected explained variances}
\label{section: projected explained variance}
We consider in this section the case where the adjusted components $Y\p$ in step 2 are obtained by projection of the components $Y$ on the chosen orthonormal basis  $X$ of $\spann\{Y\}$.
 The adjusted component $y\p_{j}$ is hence defined by~:
  \begin{equation}
\label{532}\nonumber
  {y\p}_{j} =  \langle y_{j}\,,x_{j}\rangle x_{j} %= m_{j,j}\,x_{j} 
  \quad , \quad j=1\dots\nl \ .
\end{equation}  

\smallskip
\noindent The so called \emph{ projected explained variance} of $Y$ estimated with $X$ is then defined by~:
\begin{equation}
\label{532-7}
   \varproj^{X}(Y) \egaldef  \|Y\p\|_{F}^{2}  = \sum_{j=1\dots\nl} \langle y_{j}\,,x_{j}\rangle^{2}
  % = \sum_{j=1\dots \nl} m_{j,j}^{2} 
\end{equation}
Before specifying the rule of step 1 which  associates  an orthonormal basis $X$  to the possibly correlated components $Y$, we give some properties of $\varproj^{X}(Y)$ which hold independently of the chosen basis $X$~:

\medskip

\begin{Lemma} \emph{(Projected explained variances)}
\label{lem 3}
For any unit norm loadings $Z$ and any basis $X$ chosen according to the rule \eqref{530}, the projected explained variance   of $Y=AZ$defined by \eqref{532-7}  satisfies conditions 1, 2 and 3 and~:
\begin{equation}
\label{534-2}
     \varproj ^{X}(Y)  \leq   \varsubspace(Y) \leq \varpca \ .
 \end{equation}
\end{Lemma}
\noindent The proof is in Section \ref{proof of lem 3} of the Appendix. We can now specify the rules  for the selection of the orthonormal basis $X$, which give each, according to Lemma~\ref{lem 3}, a projected explained variance satisfying conditions 1 to 3.

\paragraph{QR projected explained variance.} We choose here to associate to the components $Y$, the orthonormal basis $X=Q$ obtained by QR-decomposition of $Y$ as recalled in \eqref{531}.The vector $x_{1}$ of the basis $X=Q$ is chosen in the direction of the component with the larger norm, and the remaining components are projected on the orthogonal subspace to $x_{1}$. Then $x_{2}$ is determined by the same process applied in the orthogonal subspace, and so on. This reordering ensures that the  basis $X=Q$ associated to $Y$ will point in the direction of $Y$ primarily for the components of larger variance. The (order dependent) resulting \emph{QR~projected~explained~variance}   is given by~:
\begin{equation}
\label{534}%\nonumber
 \varprojqr(Y) =   \sum_{j=1\dots \nl} \langle y_{j}\,,x_{j}\rangle^2 = \sum_{j=1\dots \nl} r_{j,j}^{2} .
 \end{equation}
It coincides with the \emph{adjusted variance} introduced in \cite{zou2006sparse}. 

\paragraph{UP Projected explained variance.}  We choose now to associate to the components $Y$,  the orthonormal basis $X=U$ obtained by UP-decomposition (polar decomposition) of $Y$ as recalled in \eqref{531}. 
The basis $X=U$ does its best to point in the same direction as the components $Y$, in that it maximizes the scalar product $\langle Y,X \rangle_{F} = \sum_{j=1\dots \nl} \langle x_{j},y_{j}\rangle$. The (order independent) resulting 
\emph{UP projected explained variance} is given by~:
\begin{equation}
\label{536}%\nonumber
    \varprojup(Y)  = \sum_{j=1\dots \nl} \langle y_{j}\,,x_{j}\rangle^2=\sum_{j=1 \dots \nl} p_{j,j}^{2} \ .  %= \tr\{ (\diag^{2} (Y^{T}Y)^{1/2} \} \ .
\end{equation}

\smallskip

\paragraph{Optimal projected explained  variance.} 
The idea here is to associate to the components $Y$, the basis $X$ which gives the largest projected explained variance  defined in \eqref{532-7}. This choice satisfies obviously condition \eqref{530} and the so called \emph{optimal projected explained variance} is defined by~:
\begin{equation}
\label{534-10}
 \varprojopt(Y) = \max_{X^{T}X = I_{\nl}} \sum_{j=1\dots\nl} \langle y_{j}\,,x_{j}\rangle^{2} \ ,
\end{equation}
The numerical computation of the optimal projected explained variance requires the maximization
of the convex function $X \leadsto %\varproj ^{X} Y =
 \sum_{j=1\dots\nl} \langle y_{j}\,,x_{j}\rangle^{2}$ under the constraint 
$X^{T}X = I_{\nl}$. This can be done using the algorithm of \cite{JNRS2010}, which gives here~:
\begin{equation}
\label{534-11}
  X_{k+1} = \polar \big(2 \, Y \diag(X_{k}^{T}Y)    \big) \quad , \quad
  X_{0}=U=\polar(Y) \ .
\end{equation}
where $2Y \diag(X_{k}^{T}Y)$ is the gradient at $X_{k}$ of the function 
$X \leadsto \sum_{j=1\dots\nl} \langle y_{j}\,,x_{j}\rangle^{2}$.

%-----------------------

 \section{Comparison of the explained variances}
%\section{Size comparison}
\label{comparison of explained variances}

In this section, the six variances explained by non orthogonal components (see Table \ref{res_def_var}), are compared theoretically and numerically. 

 \begin{table}[ht]
\centering
\begin{tabular}{llcl}
  \hline
 Name & Notation & Definition & Short name \\ 
  \hline
subspace variance & $ \varsubspace$ & [\ref{540}] & subspVar \\ 
QR normalized variance & $\varnormqr$ & [\ref{539-0}] & QRnormVar  \\ 
UP normalized variance & $\varnormup$ & [\ref{539-1}]  & UPnormVar \\ 
 QR projected variance &  $\varprojqr$ & [\ref{534}]& QRprojVar  \\ 
 UP projected variance & $\varprojup$  & [\ref{536}] & UPprojVar \\ 
 optimal projected  variance &  $\varprojopt $ & [\ref{534-10}] & optprojVar \\ 
   \hline
\end{tabular}
\captionof{table}{Summary of for the 6 variance definitions.}
\label{res_def_var}
\end{table}

\subsection{What we know}
\label{ss-size-theoretical-results}

\begin{figure}[h]
\begin{center}
\centerline{\resizebox{28em}{!}{\includegraphics{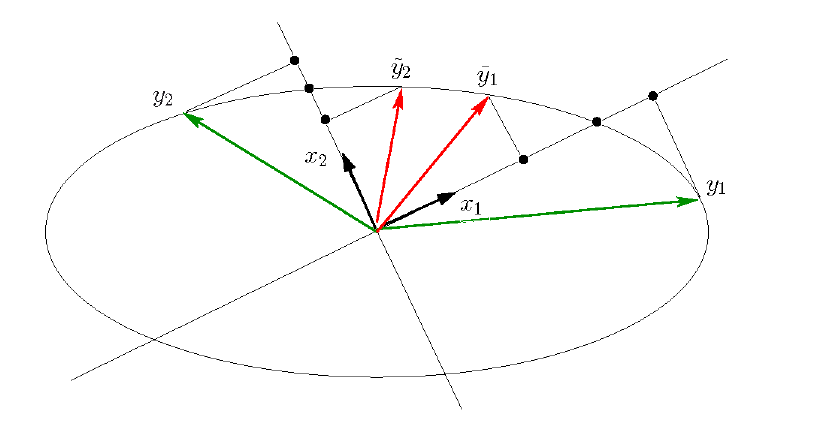}}}
\caption{The two sets of components $Y=[y_{1}, y_{2}]$ and $\tilde{Y}=[\tilde{y}_{1}, \tilde{y}_{2}]$ have been chosen such that their polar decomposition produces the same basis $X=[x_{1}, x_{2}]$, and one sees that~:
 $ \varprojup(\tilde{Y}) \leq \varnormup(\tilde{Y}) = \varnormup(Y) \leq  \varprojup(Y) $.}
\label{fig 1}
\end{center}
\end{figure}

We give first theoretical results on the relative magnitudes of the 6 explained variances~:

\begin{itemize}
  \item[-]  The subspace explained variance is larger than any of the five other variances 
   (Lemmas  \ref{lem 3} and \ref{lem 4}). It is even larger than expected when components or loadings are orthogonal without being left and right singular vectors (see \eqref{546-1} and  \eqref{546-3}).
  \smallskip
  \item[-] The optimal projected variance is greater by definition than  any other projected variance, in particular greater than the QR and the UP projected variance. 
  \smallskip
  \item[-] There is no natural order between the QR and UP projected variances~: when the components $Y$ are of equal norm, the basis $X$ which maximizes 
  $\varproj^{X}(Y)$ is $\polar(Y)$, which implies in particular that~:
 \begin{equation}
\label{539-27}
\varprojopt(Y) =\varprojup(Y) \geq \varprojqr(Y) \ .
\end{equation}
But the converse of the last inequality can hold when the norms of the components are very different~: for $\nl = 2$, one checks that $\|y_{2}\|/\|y_{1}\|$ small enough implies that $\varprojup(Y) \leq \varprojqr(Y)$.
 \smallskip
  \item[-]  There is no natural order between the QR and UP normalized variances~: for components $Y$ such that the basis $X$ associated by QR-decomposition coincides with the $\nl$-first left singular vectors $\V_{\nl}$ of $A$, one has, according to Lemma \ref{lem 4}~:
 \begin{equation}
\label{539-3}\nonumber
  \varnormqr(Y) = \sum_{j=1\dots \nl}\sigma_{j}^{2} \geq \varnormup(Y) \ ,
\end{equation}
  with a strict inequality as soon as $Y$ and $\V_{\nl}\diag\{ \sigma_{j}\}$ don't coincide.
 The same reasoning with the polar decomposition in place of the QR decomposition shows that the converse inequality can happen.
 \smallskip
\item[-] There is no natural order between the variances defined by projection and normalization, as illustrated in Figure  \ref{fig 1}.
\end{itemize}

 \subsection{What we see}
 \label{size-numerical-experiments}

We  compare now  numerically the six explained variances. The comparison  is made on non orthogonal components $Y$ obtained by applying sparse PCA to  simulated data matrices $A$. The matrices $A$ are obtained using the simulation scheme of \cite{chavent2023-cp-two} based on $m=4$ underlying loadings vectors of size $p=20$ and $m=4$ underlying first eigenvalues which are chosen to be either close or different.  More precisely,  $100$  matrices $A$ of size $n \times 20$ were drawn randomly using either the  ``close eigenvalues'' or the ``different eigenvalues" scheme. Three sparse PCA algorithms  \citep[see][]{chavent2023-cp-two} were applied to each matrix $A$ for a grid of 101 values of sparsity parameters $ \lambda \in [ 0,1]$. Finaly, $30300=100*3*101$ loadings matrices $Z$ and components matrices  $Y=AZ$ where obtained for the ``close eigenvalues'' and for the ``different eigenvalues" scheme. The  variance explained by these components where performed using the six variance definitions  to finally obtain the  {dimensionless} {\em proportion of explained variance} (pev) defined by~:
\begin{equation}
\label{nr25}\nonumber
 pev = \frac{\var(Y) }{ \|A\|_{F}^{2}} \leq
  \frac{ \varpca}{ \|A\|_{F}^{2} } \ ,
\end{equation}
where the right inequality follows from \eqref{510a} in Condition 2, which is satisfied by all definitions.

Figure \ref{fig 8} gives  the  mean $pev$  for the 300 non orthogonal components (for the ``close eigenvalues'' and the ``different eigenvalues'' case) as a function of the sparsity parameter $\lambda$ and for each definition of explained variance. 

\begin{minipage}[c]{1\textwidth}
\centering
\resizebox{22em}{!}{\includegraphics{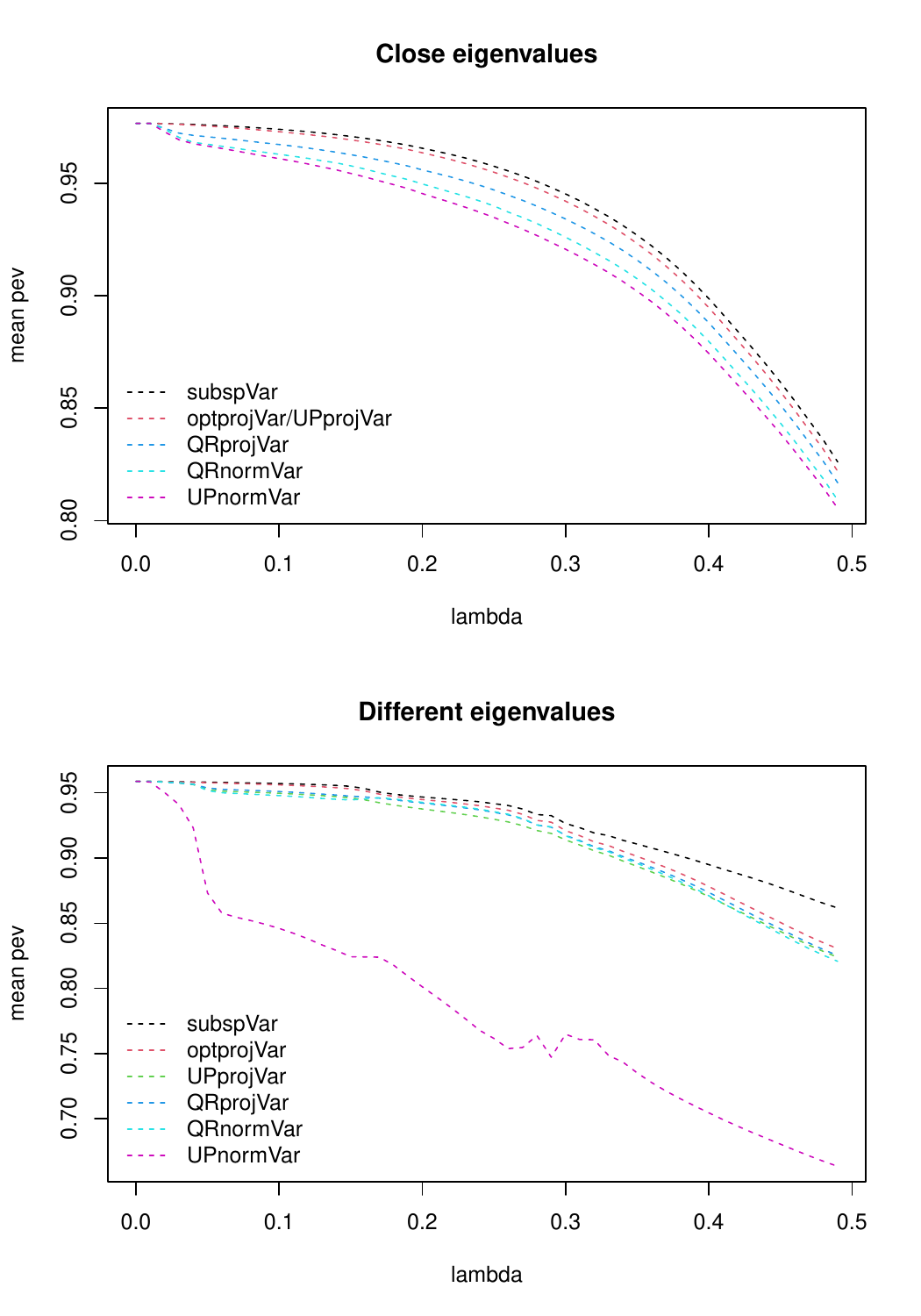}}
\captionof{figure}{Comparison of the  mean pev (proportion of explained variance) over two sets of components (top and bottom) as function of  sparsity parameter $\lambda$ for the six variance definitions.}
\label{fig 8}
\end{minipage}

\noindent For the ``close eigenvalues'' case (top), the six definitions give relatively close results and the results produced by optprojVar and UPprojVar are so close that they cannot be distinguished on the figure (remember that they would coincide were the norms equal, see \eqref{539-27}). One sees also that  the results seem to be in a certain order for all~$\lambda$~:
\begin{equation}
\label{nr 30}
\nonumber
\text{subspVar  $\geq$ optprojVar $\geq$ UPprojVar $\geq$ 
QRprojVar $\geq$ QRnormVar $\geq$ UPnormVar}
\end{equation}
(the two first inequalities are not a surprise, as they hold theoretically). For the ``different eigenvalues'' case (bottom), a zoom on the curves shows that subspVar and optprojVar are again larger than all other variances. But the previous apparent order between UPprojVar, QRprojVar and QRnormVar is not longer observed. The UPnormvar remains the smallest but its behavior seems disturbed as the sparsity paramter $\lambda$ increases.

\vspace{2ex}

Table \ref{tab_sd_var} shows that all definitions but UPnormVar (in the ``different eigenvalues'' case) exhibit quite similar dispersions over the 300 realizations (for $\lambda=0.3$).

\begin{table}[ht]
\centering
\begin{tabular}{rrr}
  \hline
 & Close eigenvalues & Different eigenvalues \\ 
  \hline
  subspVar & 0.63 & 1.63 \\ 
   optprojVar & 0.66 & 1.38 \\ 
  UPprojVar & 0.66 & 1.06 \\ 
  QRprojVar & 0.72 & 1.25 \\ 
  QRnormVar & 1.06 & 1.33 \\ 
  UPnormVar & 1.21 & 7.74 \\ 
   \hline
\end{tabular}
\captionof{table}{Standard deviations$\times 100$  of the six pev (proportion of explained variance) obtained for $\lambda=0.3$ with the three algorithms over the two sets of components (close eigenvalues and different eigenvalues).}
\label{tab_sd_var}

\end{table}

\subsection{Ranking properties}
\label{ranking properties of explained variances}

The proportions of explained variance are meant to be used for the ranking of algorithms, so it is important to figure out wether or not definitions $i$ and $j$  of explained variance will rank in the same order the components $Y$ and $Y^{\prime}$ obtained from possibly different algorithms and/or sparsity parameter $\lambda$ and/or realization of the data matrix $A$. The components obtained with the 3 algorithms, the 50 smallest values $\lambda$ and 100 realizations of $A$ (``different eigenvalues '' case) gave $15000\times14999/2$ couples $(Y,Y^{\prime})$ to be tested. Among these couples, we may consider as  
$\epsilon$-distinguishable from the point of view of our explained variances those for which
\begin{equation}
\label{nr30}\nonumber
 |\pev_{i}(Y)-\pev_{i}(Y^{\prime})| \geq \epsilon  \quad  \mbox{ for all } \quad i=1 \dots 6 
\end{equation}
for some $\epsilon\geq 0$. Table \ref{tab 2} shows the percentage of cases where $\pev_{i}$ and $\pev_{j}$ rank identically components $Y$ and $Y^{\prime}$ among all $\epsilon$-distinguishable couples. 

\begin{table}[ht]
\centering
\begin{tabular}{rrrrrr}
  \hline
 & optprojVar & UPprojVar & QRprojVar & QRnormVar & UPnormVar \\ 
  \hline
subspVar &   79.98 & 71.05 & 70.71 & 69.75 & 56.19 \\ 
  optprojVar &    & 88.93 & 89.87 & 89.19 & 73.71 \\ 
  UPprojVar &    &  & 96.22 & 95.27 & 84.62 \\ 
  QRprojVar &    &  &  & 98.34 & 83.15 \\ 
  QRnormVar &    &  &  &  & 82.75 \\ 
  UPnormVar &    &  &  &  &  \\ 
   \hline
\end{tabular}

\vspace{2ex}

% eps = 0.001
\centering
\begin{tabular}{rrrrrr}
  \hline
 & optprojVar & UPprojVar & QRprojVar & QRnormVar & UPnormVar \\ 
  \hline
subspVar   & 86.13 & 83.78 & 84.70 & 84.57 & 68.68 \\ 
 optprojVar  &  & 96.40 & 98.57 & 98.32 & 81.14 \\ 
  UPprojVar   &  &  & 97.81 & 97.72 & 84.74 \\ 
  QRprojVar   &  &  &  & 99.66 & 82.55 \\ 
  QRnormVar  &  &  &  &  & 82.66 \\ 
  UPnormVar   &  &  &  &  &  \\ 
   \hline
\end{tabular}

\vspace{2ex}

% eps = 0.01
\centering
\begin{tabular}{rrrrrr}
  \hline
& optprojVar & UPprojVar & QRprojVar & QRnormVar & UPnormVar \\ 
  \hline
subspVar   & 89.57 & 89.57 & 89.57 & 89.57 & 68.80 \\ 
  optprojVar   &  & 100.00 & 100.00 & 100.00 & 79.23 \\ 
  UPprojVar   &  &  & 100.00 & 100.00 & 79.23 \\ 
  QRprojVar   &  &  &  & 100.00 & 79.23 \\ 
  QRnormVar   &  &  &  &  & 79.23 \\ 
  UPnormVar   &  &  &  &  &  \\ 
   \hline
\end{tabular}

\captionof{table}{The  entry of each table on line $i$ and  column $j$ gives the percentage of $\epsilon$-distinguishable couples $Y,Y^{\prime}$ which are ranked identically by $\pev_{i}$ and $\pev_{j}$. Top~: $\epsilon=0$, middle~: $\epsilon=10^{-3}$, bottom~: $\epsilon=10^{-2}$.}
\label{tab 2}
\end{table}

The good news is that all three projected variances optprojVar, UPpojVar and QRprojVar, as well as the normalized variance QRnormVar,  produce the same ranking as soon as one considers that differences in proportion of explained variance  under $10^{-2}$ are not significative. For the same $\epsilon$, the two other definitions subspVar and UPnormVar still produce quite different rankings.  The numerical results of this section 
%\ref{ranking properties of explained variances} 
show that all investigated definitions rank the explained variance of components in essentially the same order, and hence can all be used to compare the variance explained by components obtained by different algorithms or with different parameters.
Of course, this is only an experimental result based on our simulated data sets, that needs to be confirmed by further numerical tests.

\section{Explained variance block PCA formulations}
\label{explained variance block pca formulations}

With the objective in mind to get rid of the orthogonality constraints in the usual block PCA formulation  \eqref{583}, we discuss in this section the possibility of using the above defined explained variance measures as objective function 
in new block PCA formulations where the loadings are subject to unit norm constraints only~:
\begin{equation}
\label{105-3}
 \max_{\text{\scriptsize{
 $
 %\begin{array}{c}
  \|z_{j}\|=1   ,
  j=1 \dots \nl
%\end{array}
$ } }}  \var (AZ) \ \ =
  \varpca \ ,
\end{equation}
As seen in Lemmas \ref{lem 1}, 
\ref{lem 4} and \ref{lem 3}, property \eqref{105-3} holds 
for the subspace variance and the five normalized and projected  variance definitions. 
However, from these six definitions, only the three projected explained variances satisfy (Lemma \ref{lem 3}) all necessary conditions 1, 2 and 3 of section  \ref{how to define the explained variance}  required for a proper definition of explained variance. So we shall limit our search for an explained variance block formulation to these three definitions.

A projected explained variance $\varproj(AZ)$ will provide a block PCA formulation  \eqref{105-3} if and only if
its  sole maximizers are the loadings $Z$  made of the $\nl$ first right singular vectors 
$\W_{\nl}$ in any order, up to a $\pm 1$ multiplication, or in short~:
\begin{equation}
\label{evb condition}
Z \egalperm \W_{\nl} \text{ are the sole maximizers of $\var(AZ)$} \ .
\end{equation}
We denote also by 
\begin{equation}
\label{definition Am}
A_{\nl} \egaldef \V_{\nl}\Sigma_{\nl}\W_{\nl}^{T} \quad \text{ with } \quad
 \Sigma_{\nl} \egaldef \diag\{\sigma_{1} \dots \sigma_{\nl}\}
\end{equation}
 the restriction of the matrix $A$ to the subspace of the $\nl$ first right singular vectors $\W_{1}\dots \W_{\nl}$. 
 It is a bijection from the right to the left singular subspaces associated to the $\nl \leq r$ first singular values.  In the singular basis $\V_{\nl},\W_{\nl}$, \,
  $A_{\nl}$ reduces to $\Sigma_{\nl}$. 

\medskip

The next lemma characterizes the maximizers of any projected variance~:

\medskip

 \begin{Lemma}{Maximizers of projected variances}
 \label{lem 3-2}
  \begin{enumerate}
 \item  For a given set of components $Y$,  the basis $X$ %of $\spann\{Y\}$ 
 which gives the largest projected explained variance $\varproj^{X}(Y)$ is necessarily a solution of~:
   \begin{equation}
\label{534-5}
   Y \diag(X^{T}Y) = X P \text{ with $P^{T} \!= P$ and $P \geq 0$ }
\end{equation}
\item
 The projected explained variance $\varproj^{X}(AZ)$ achieves it PCA maximum value $\varpca$ if and only if $Z$ and $X$ satisfy the three following conditions~:

\begin{eqnarray}
\label{534-41}
& \spann\{Z\} = \spann\{\W_{\nl}\} \\
%\text{$Z$ is a  (non necessarily orthogonal) unit norm basis of $\W_{\nl}$}& \\
\label{534-42}
 & \text{the loadings $Z$ are 
  $ %\langle \bullet,\bullet\rangle_{ (A^{T}_{\nl}A_{\nl})^{-1}} =
  (A^{T}_{\nl}A_{\nl})^{-1}$-- orthogonal} & \\
\label{534-53}
& \text{the basis $X$ associated to $Y=AZ$ is } 
X=[n_{1} \dots n_{\nl}]  \ . &
\end{eqnarray}
%where $A_{\nl}$ is the restriction of $A$ to $\W_{\nl}$, and where
%where $A_{\nl}$ is the matrix $A$ restricted to $\W_{\nl}$, and 
where $n_{1} \dots n_{\nl}$ denote unit normals at $y_{1}\dots y_{\nl}$ to the ellipsoid $\E^{\nl}$ of $\spann\{\V_{\nl}\}
%=\spann\{Y\}$
$
 image by $A_{\nl}$ of the unit sphere of $\spann\{\W_{\nl}\}$  (see Figure \ref{fig 0})
\end{enumerate}
\end{Lemma}
The proof is in Appendix  \ref{proof of lem 3-2}
\begin{figure}[h]
\begin{center}
%\centerline{\resizebox{28em}{!}{\includegraphics{20221217_Figure_manuscrite_1}}}
\centerline{\resizebox{28em}{!}{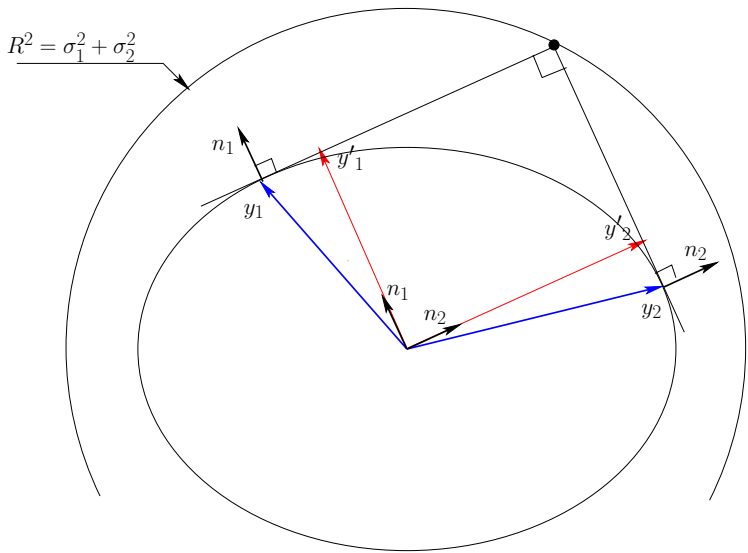}}
\caption{Illustration of condition \eqref{534-53} for $\nl =2$. 
The ellipse $\E^{2}$ is the image by $A$ of all unit norm loadings $z$ of the right singular space $\W_{2}$, with half axes $\sigma_{1}$ and $\sigma_{2}$. Let $y_{1},y_{2}$ be two (possibly correlated) components such that the normals  $n_{1}, n_{2}$ to $\E^{2}$ at  $y_{1},y_{2}$ are orthogonal, the tangents are orthogonal too and hence meet on the Cartan Circle of radius $R=(\sigma_{1}^{2}+\sigma_{2}^{2})^{1/2}$.  Then for the choice $X=(n_{1}, n_{2})$ one has
$\varproj^{X} Y = \|{y\p}_{1}\|^{2}+\|{y\p}_{2}\|^{2}=R^{2}=\sigma_{1}^{2}+\sigma_{2}^{2}$.
%The proof is in section \ref{proof of lem 3-2} of Appendix.
}
\label{fig 0}
\end{center}
\end{figure}

Condition \eqref{534-42} reduces, in the singular basis $\W_{\nl}$, to $\Sigma_{\nl}^{-2}$--orthogonality of the loadings. It ensures 
%Condition \eqref{534-42} ensures 
that the normals $n_{j}$ are orthogonal, and hence can be chosen as the basis $X$ associated to $Y$  in \eqref{534-53}.
Of course, the SVD solutions $Z  \egalperm  \W_{\nl}$  of PCA  satisfy always  \eqref{534-41} \eqref{534-42} \eqref{534-53}.

Once a rule $Y \leadsto X$ has been specified, 
%conditions \eqref{534-41} \eqref{534-42} \eqref{534-53} 
points 1 and 2 of the lemma
will make it possible to determine wether or not  
\eqref{evb condition} is satisfied.
% the PCA solution $Z  \egalperm  \W_{\nl}$ is the unique maximizer of the associated projected explained variance.

 \medskip
 
 \subsection{Maximizers of $\varprojqr(Y)$}
 %\paragraph{Maximizers of $\varprojqr(Y)$}
 
 The projected explained variance  $\varprojqr(AZ)$ attains its maximum
 $\varpca$ if and only if~:
 %Its maximizers are given by~:
\begin{equation}
\label{qr pca formulation}
  %\varprojqr (AZ) =  \varpca = \max   \quad \Longleftrightarrow  \quad 
  Z \egalperm \W_{\nl} \ ,
\end{equation}
 When the maximum norm selection procedure is applied at each step and the components renumbered accordingly, the unique maximizer is $Z = \W_{\nl}$.
\smallskip

\noindent \textbf{Proof:} The ``if'' part of \eqref{qr pca formulation} is trivial, we prove the ``only if'' part~: let $Y$ be such that $\varprojqr Y= \varpca
%\sum_{j=1\dot \nl} \sigma_{j}^{2}
$, and X be given by its QR decomposition $Y=XR$ with $R$ upper triangular. 
By hypothesis, $X$ maximizes 
$\var^{X}\,Y$, and point 1 of Lemmas~\ref{lem 3-2} implies that
$Y \diag{(X^{T}Y)} =XP= X R \diag{(X^{T}Y)} $, with $P=R \diag{(X^{T}Y)}$ symmetric positive. This implies that $R$ is a diagonal matrix, so that all $x_{j}$ point in the direction of $y_{j}$. But 
 property 2 of Lemma \ref{lem 3-2} shows that the $x_{j}$'s are also normal to the 
ellipsoid $\E^{\nl}$ at $y_{j}$, which can happen only if $y_{j}=Az_{j}$ coincides with its principal axes and hence each $z_{j}$ is one of the $\nl$ first right eigenvectors $v_{j}$. \cqfd

\medskip

\begin{Proposition}
\label{qr block pca}
The QR projected explained variance $\varprojqr(AZ)$
(the \emph{adjusted variance} of  \cite{zou2006sparse}) associated to unit norm loadings $Z$ provides a block PCA formulation~:
\begin{equation}
\label{qr block optimisation}
 %\max_{Z\in \bns }  \var (AZ) \ \ 
 \max_{\text{\scriptsize{
 $
 %\begin{array}{c}
  \|z_{j}\|=1  ,
  j=1 \dots \nl
%\end{array}
$ } }}  \varprojqr (AZ) \ \ =
  % \sum_{j=1 \dots \nl} \sigma_{j}^{2}  \ ,
  \varpca \ ,
\end{equation}
which admits the SVD solution \eqref{qr pca formulation} as ``unique'' maximizer.
\end{Proposition}

\smallskip

\noindent  Numerical implementation of this formulation requires 
 the computation of the gradient of the $Z \leadsto  \varprojqr(AZ)$ function, which is defined through the QR decomposition of $Y=AZ$. 
 This can be done by the adjoint state method \cite{cha:nllsip2010}, which is feasible but may be cumbersome. The block PCA formulation \eqref{qr block optimisation}  can be used as starting point for the design of sparse PCA algorithms, keeping in mind that enforcing sparsity by subtracting  the $\ell^{1}$ norm of loadings leads to a difficult, though tractable, non smooth optimization problem.

 \subsection{Maximizers of $\varprojup(Y)$}
 %\paragraph{Maximizers of $\varprojup(Y)$}
 
\begin{figure}[h]
\begin{center}
%\centerline{\resizebox{28em}{!}{\includegraphics{20221217_Figure_manuscrite_2}}}
\centerline{\resizebox{28em}{!}{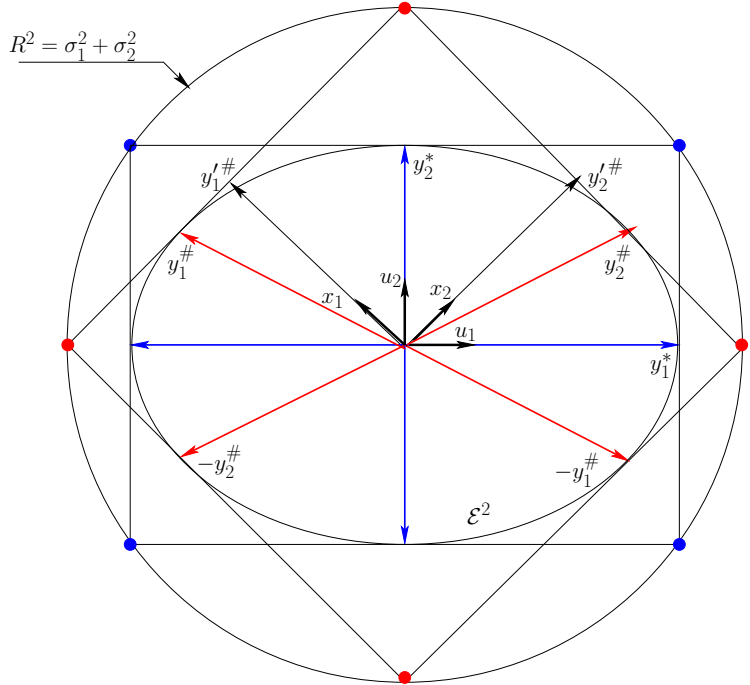}}
\caption{Illustration, for $\nl =2$, of the parasitic maximizer $Y^{\#}$ of $\varprojup$. The polar decomposition associates to components $Y^{\#} = y^{\#}_{1},y^{\#}_{2}$ orthonormal vectors $X=x_{1},x_{2}$
  such that $\|{y_{1}\p}^{\#}\|=\|{y_{2}\p}^{\#}\|$.
 The red points on the Cartan circle correspond to the parasitic maximizers 
 $Y^{\#} = \pm y^{\#}_{1},\pm y^{\#}_{2}$, the blue points to the SVD solution $Y^{*} =
  \pm y^{*}_{1},\pm y^{*}_{2} = \pm \sigma_{1}u_{1},\pm \sigma_{2}u_{2}$ }
\label{fig 0a}
\end{center}
\end{figure}

 The projected explained variance $\varprojup(AZ)$ attains its maximum $\varpca$ if and only if~:
\begin{equation}
\label{up proj pca formulation}
   %\varprojup (AZ) =  \varpca = \max   \quad \Longleftrightarrow
   %\left \{ \begin{array}{l}
   Z \egalperm \W_{\nl} \quad \text{(SVD solution)}  
   \quad \text{or} \quad 
   Z=Z^{\#}
%\end{array} \right .
\end{equation}
 where the ``parasitic'' solution $Z^{\#}$ is such that the components $Y^{\#}=AZ^{\#}$ satisfy 
 $\langle y^{\#}_{1},x_{1}\rangle = \dots = \langle y^{\#}_{\nl}, x_{\nl}\rangle $,
 with the hyperplanes tangent to $\E^{\nl}$ at $y^{\#}_{j}$ delimiting an 
 $\nl$-dimensional orthant  of $\spann\{ \W_{\nl}\}$. For $\nl
= 2$, the parasitic solution is illustrated on figure \ref{fig 0a}, where one sees that the components $y^{\#}_{j}$ (in red) corresponding to the choice of different principal axes for the intersection of the tangents coincide up to a multiplication by~$\pm 1$; the SVD components are in blue. The proof is in Appendix \ref{proof of property up proj pca formulation}.

\smallskip

So $\varprojup(AZ)$ cannot be used for the construction of a block PCA formulation like \eqref{105-3}, as the optimization algorithm might converge to the parasitic solution $Z^{\#}$~!

 \subsection{Maximizers of $\varprojopt(Y)$}
 %\paragraph{Maximizers of $\varprojopt(Y)$}
 
 According to point 2 of Lemma \ref{lem 3-2}, the optimal projected explained variance $\varprojopt(AZ)$ attains its maximum $\varpca$ if and only if~:
\begin{equation}
\label{534-14}
%\hspace{-1em} \varprojopt (AZ) \! = \!\! \varpca = \ \max \Longleftrightarrow
%\left \{ \begin{array}{l}
   \spann\{ Z \} = \spann\{ \W_{\nl} \} 
   \quad \text{ and } \quad
    \text{the loadings $Z$ are $(A^{T}_{\nl}A_{\nl})^{-1}$-- orthogonal} \ .
   %\langle z_{j}, (A^{T}_{\nl}A_{\nl})^{-1} z_{k}\rangle_{\R^{\nvs}} =  0   \ , \  j \neq k 
%\end{array} \right. ,
% \spann Z = \spann \W_{\nl} \  \text{and}  \  
% \langle z_{j},z_{k}\rangle_{ (A^{T}_{\nl}A_{\nl})^{-1} } =  0  % \text{ for } 
%  \ , \  j \neq k \ .
\end{equation}
%and hence admits a continuum of maximizers.
 This situation is similar to the maximization \eqref{583}  of 
$\|AZ\|^{2}$ under orthonormality constraints $Z^{T}Z=I_{\nl}$, were the maximum is attained for all orthonormal $Z$ which span $\W_{\nl}$. But the difference is that the orthogonality condition \eqref{534-14}-right is not a constraint for the maximization of $\varprojopt(AZ)$, it just happens to be satisfied by the maximizer
%$(A^{T}_{\nl}A_{\nl})^{-1}$-orthogonality 
~!

\subsection{Weighted optimal projected explained variance}
\label{weighted projected explained variance}

In order to select the SVD solution $Z = \W_{\nl}$ among the maximizers of 
$\varprojopt(AZ)$, 
%we proceed similarly to the case  of the usual block PCA formulation   \eqref{583}~: 
we introduce  weights $\mu_{j}$ such that~:

\begin{equation}
\label{136}
  \mu_{1} \geq  \mu_{2} \geq \dots  \mu_{\nl} >0 \ ,
 \end{equation}
and define a  \emph{weighted optimal projected variance} by~:
 \begin{equation}
\label{136-4}\nonumber
  \varprojoptmu (AZ)=   \max_{X^{T}X=I_{\nl}} \sum_{j=1\dots\nl} \mu_{j}^{2} \langle Az_{j}\,,x_{j}\rangle^{2} \ ,
\end{equation}
which coincides with $\varprojopt(AZ)$ when $\mu_{j}=1 \text{ for all } j$.  This leads to the \emph{weighted optimal projected explained variance } block PCA formulation~:
\begin{equation} 
\label{132-5}
\hspace{-0,5em}
\max_{
\text{\scriptsize{$\begin{array}{c}
  \|z_{j}\|=1    \\
  j=1 \dots \nl
\end{array}$ } }
} \hspace{-1em} \varprojoptmu(AZ)  =  \hspace{-1em}
\max_{\text{\scriptsize{
$\begin{array}{c}
  \|z_{j}\|=1    \\
  j=1 \dots \nl
\end{array}$ } }}   \hspace{-0,8em} \max_{X^{T}X=I_{\nl}} \sum_{j=1\dots\nl} \mu_{j}^{2} \langle Az_{j}\,,x_{j}\rangle^{2}  =  \hspace{-1em}
\sum_{j=1\dots\nl} \mu_{j}^{2} \sigma_{j}^{2} \ . \hspace{-1,3em}
\end{equation}
The nice properties  of this formulation are recalled in the next proposition~:
\smallskip
\begin{Proposition}
\label{pro 1-Z}
Let the singular values of $A$ satisfy~:
\begin{equation}
\label{137}\nonumber
  \sigma_{1}>\sigma_{2}> \dots >\sigma_{\nl}>0 \ ,
\end{equation}
and the weights $\mu_{j}$ satisfy (\ref{136}).  Then the PCA loadings $Z=\W_{\nl}$ and normalized components $X=\V_{\nl}$ defined in  \eqref{110} are \emph{one} solution of the block PCA formulation  \eqref{132-5} when the weights $\mu_{j}$ are constant, and its \emph{ unique solution} (up to a multiplication by $\pm1$ of each column of course) when the weights $\mu_{j}$ are strictly decreasing,
in which case 
the maximizers $Z^{*}$ and $X^{*}$ are independent of the weights $\mu_{j}$. The  (unweighted) optimal projected variance explained by $Y^{*}=AZ^{*} $ is~: 
\begin{equation}
\label{147a}\nonumber
   \varprojopt \, (Y^{*})   =  \varpca  \leq 
   \|A\|_{F}^{2} \ .
\end{equation}
%with $X^{*}$ and $Z^{*}$ related by~:
% \begin{equation}
%\label{162}\nonumber
%  x_{j}^{*}=(Az_{j}^{*}) / \|Az_{j}^{*}\| \quad , \quad  z_{j}^{*}=(A^{T}x_{j}^{*}) / \|A^{T}x_{j}^{*}\| 
%  \quad \mbox{for} \quad j=1 \dots \nl \ .
%\end{equation}
\end{Proposition}
\noindent\textbf{Proof:} Exchanging the order of maximization in the center term of \eqref{132-5} solving analytically the maximization with respect to $Z$ gives~:
\begin{equation} 
\label{132-8}
\max_{\text{\scriptsize{$\begin{array}{c}
  \|z_{j}\|=1    \\
  j=1 \dots \nl
\end{array}$ } }} \varprojoptmu(AZ)  
 %\max_{Z \in \bpnvs}   \max_{X\in\stns} \sum_{j=1\dots\nl} \mu_{j}^{2} \langle Az_{j}\,,x_{j}\rangle^{2}  
 = \max_{X^{T}X=I_{\nl}}  \sum_{j=1\dots\nl} \mu_{j}^{2} \|A^{T}x_{j}\|^{2} \ .
\end{equation}
The last term in resp. \eqref{132-8}
is the maximization of a weighted Rayleigh quotient
for $A^{T}$, which is known to be equivalent to a PCA problem for $A^{T}$, and hence for $A$ (see for example \cite{AMS2008}, recalled as Theorem~\ref{thm GRQ} in the Appendix  for the case of constant weights, and 
\cite{Brockett1991} for  the case of decreasing weights).
\cqfd

%From a numerical point of view, maximizing $\varprojoptmu$ amounts hence to maximize the convex function $X \leadsto 
%\sum_{j=1\dots\nl} \mu_{j}^{2} \|A^{T}x_{j}\|^{2}$ under the constraint
%$X^{T}X=~I_{\nl}$, for example by the algorithm   \cite{JNRS2010}. 
Of course, formulation \eqref{132-5} is of little interest for PCA itself, as there exists plenty of other efficient solution methods. But it will provide the starting point for the design of a sparse PCA algorithm, to be developped in the companion paper  \cite{chavent2023-cp-two}.

\section{Conclusion}

We have investigated the problem of defining the part  of the variance of a data matrix explained by correlated components, such as those which arise when sparse loadings are searched for. We have established three  compatibility conditions to be satisfied by any such explained variance definition in order to be
compatible with the Principal Component Analysis (Condition 1), and to ensure a loss in explained variance when the components are correlated (Conditions 2 and 3). We have proved that the two existing and the four new definitions~:

-  Subspace %explained variance 
(total variance of  \cite{SH2008}),

- QR normalized, %explained variance

- UP or Polar normalized, %explained variance

-  QR projected %explained variance 
(adjusted variance of  \cite{zou2006sparse}),

- UP or Polar projected, %explained variance

- Optimal projected %explained variance

\noindent
all satisfy the two first compatibility conditions, but that only the three projected explained variance satisfy also the third one and provide proper explained variance definitions.

%but the Subspace explained variance, which can be too large in specific situations, satisfy the three compatibility conditions.

Numerical experiments have shown that the choice of a specific definition for
the ranking of correlated components by explained variance is not critical. But we have shown that only the QR  and the (weighted) optimal projected explained variance definitions admit the SVD solution as unique maximizer, and hence provide new explained variance block PCA formulations rid of orthogonality constraints on loadings.
Their use for the construction of a group sparse PCA algorithm is the subject of a second paper \cite{chavent2023-cp-two}.

%------------------------

\section{Appendix }
\label{appendix -1}

\subsection{Generalized Rayleigh quotient}
\label{properties of generalized rayleigh quotient}
This is a classical result, see for example \cite{AMS2008} and  \cite{Brockett1991}~:
\medskip

\begin{Theorem}
\label{thm GRQ} 
Let the loadings $Z$ satisfy~:
\begin{equation}
\label{570-1}\nonumber
  Z  =  [z_{1} \dots z_{\nl}]\in \R^{\nvs \times \nl}  \quad, \quad \rank(Z)=\nl \leq \rank(A) \egaldef r  \ .
\end{equation}
Then the generalized Rayleigh quotient
\begin{equation}
\label{570-0}\nonumber
  \tr\{ (Z^{T}A^{T}AZ)(Z^{T}Z)^{-1}\}
\end{equation}
 satisfies~:
$$
  \tr\{ (Z^{T}A^{T}AZ)(Z^{T}Z)^{-1} \} \leq \sigma_{1}^{2}+\dots+\sigma_{\nl}^{2}\leq \|A\|_{F}^{2} \ ,
$$
and the left inequality becomes an equality if and only if~:
$$
  \spann \{Z\} = \spann \{ \ww_{1}\dots \ww_{\nl}\} \ ,
$$
where $\ww_{1}, \dots \ww_{\nl}$ are the $\nl$ first right singular vectors of $A$.

\end{Theorem}

\subsection{Proof of Lemma \ref{lem 1}}
\label{proof of lem 2}
Definition \eqref{540} of the subspace explained variance and the properties of the Rayleigh quotient  $\tr\{ Z^{T}A^{T}A Z(Z^{T}Z)^{-1})\}$ recalled in Theorem \ref{thm GRQ} show that \eqref{510} and \eqref{510a}, and hence Properties 1 and 2, hold as well as \eqref{546-2} and  \eqref{546-3}.

It remains to prove \eqref{546-1} which shows that Condition 3 does not hold. 
So let $Y=AZ$ be orthogonal components~:
 \begin{equation}
\label{900}\nonumber
  \langle y_{j},y_{k}\rangle = 0 \  , \ j,k=1 \dots \nl,j\neq k
\end{equation}
corresponding to unit norm loadings~:
\begin{equation}
\label{901}\nonumber
 \|z_{j}\| =1 \ j=1 \dots \nl \ ,
\end{equation}
and define $X,T$ by~:
\begin{equation}
\label{904}\nonumber
  x_{j}= y_{j}/\|y_{j}\| \quad , \quad t_{j}=z_{j}/\|y_{j}\| 
  \quad , \quad j=1 \dots \nl \ ,
\end{equation}
so that~:
\begin{equation}
\label{908}\nonumber
  X^{T}X = I_{\nl} \ .
\end{equation} 
 Then on one side one has~:
 \begin{equation}
\label{912}
  \|Y\|_{F}^{2} =  \sum_{j=1 \dots \nl} \|y_{j}\|^{2} 
  = \sum_{j=1 \dots \nl}1/\|t_{j}\|^{2}
  = \tr\{\diag^{-1}(T^{T}T)  \} \ ,
\end{equation}
 and on the other side, as $Y$ and $X$ span the same subspace~:
\begin{equation}
\label{914}
  \varsubspace(Y) = \varsubspace(X) 
  =  \tr\{ (X^{T}X)(T^{T}T)^{-1} \} 
  =  \tr\{(T^{T}T)^{-1} \} 
\end{equation}
 Formula \eqref{546-1} will be proved if we show that~:
 \begin{equation}
\label{916}
  \tr\{\diag^{-1}(T^{T}T)  \}  \leq  \tr\{(T^{T}T)^{-1} \} \ .
\end{equation}
We use for that an idea taken from \cite{M1969}, and perform
 a QR-decomposition of $T$.
By construction, the diagonal elements of $R$ satisfy~:
\begin{equation}
\label{918}\nonumber
   0 < r_{i,i} \leq \|t_{i}\| \ .
\end{equation}
Then~:
\begin{eqnarray}
\label{920}\nonumber
  T^{T}T & = & R^{T}   Q^{T} Q\, R 
            =  R^{T}  R  \ ,\\
    \nonumber
    (T^{T}T)^{-1} & = & R^{-1} (R^{T})^{-1}  =
    R^{-1} (R^{-1})^{T} \ ,         
\end{eqnarray}
where $ R^{-1}$ satisfies~:
\begin{equation}
\label{922}\nonumber
   R^{-1} = \mbox{upper triangular matrix} \quad , \quad
   [R^{-1}]_{i,i} = 1/ r_{i,i} \ .
\end{equation}
Hence  the diagonal element of $(T^{T}T)^{-1}$ are given by~:~:
\begin{eqnarray}
\label{924-1}\nonumber
  \big[(T^{T}T)^{-1}\big]_{i,i} & = & \big[ R^{-1} (R^{-1})^{T} \big]_{i,i}  \\
  \nonumber
      & = &  [R^{-1}]_{i,i}^{2} + \sum_{j>i} [R^{-1}]_{i,j}^{2} \\
      \label{924}
      & \geq &  [R^{-1}]_{i,i}^{2}  = 1/r_{i,i}^{2}  \geq 1/\|t_{i}\|^{2} \ .
\end{eqnarray}
which gives (\ref{916}) by summation over $i=1 \dots \nl$, and \eqref{546-1} is proved.

We suppose now that  the orthogonal components $y_{j},j=1\dots \nl$ satisfy 
$ \|Y\|_{F}^{2} = \varsubspace(Y)$.
Then \eqref {912} \eqref{914} imply that equality holds in \eqref{916} and hence all inequality in \eqref{924} are equalities~:
\begin{enumerate}
  \item first inequality~: $ [R^{-1}]_{i,j}^{2}=0 \text{ for all } j>i \quad \Rightarrow  \quad
   \text{ $R^{-1}$ and hence $R$ are diagonal } $
  \item second inequality~: $1/r_{i,i}^{2}  = 1/\|t_{i}\|^{2} \quad \Rightarrow \quad
  \text{$R$ is diagonal}$
\end{enumerate} 
But $R$ diagonal implies that the $t_{j}$ - and hence also the loadings $z_{j}$ - are orthogonal, which together with the hypothesis of orthogonal components $y_{j}$, implies that $(y_{j}/\|y_{j}\|,z_{j})$ are pairs of singular vectors of $A$, which proves that $Z=\W_{\nl}$ and ends the proof of \eqref{546-1}.
 %\cqfd

\subsection{Proof of Lemma \ref{lem 3}}  
\label{proof of lem 3}

By construction, $\varproj^{X}(AZ)$ satisfies clearly conditions 1 and 3 of Section \ref{adjusted optimal and orthogonal variances}. We prove now that it satisfies moreover \eqref{534-2}, and hence also condition 2.

\begin{figure}[h]
\begin{center}
\centerline{\resizebox{28em}{!}{\includegraphics{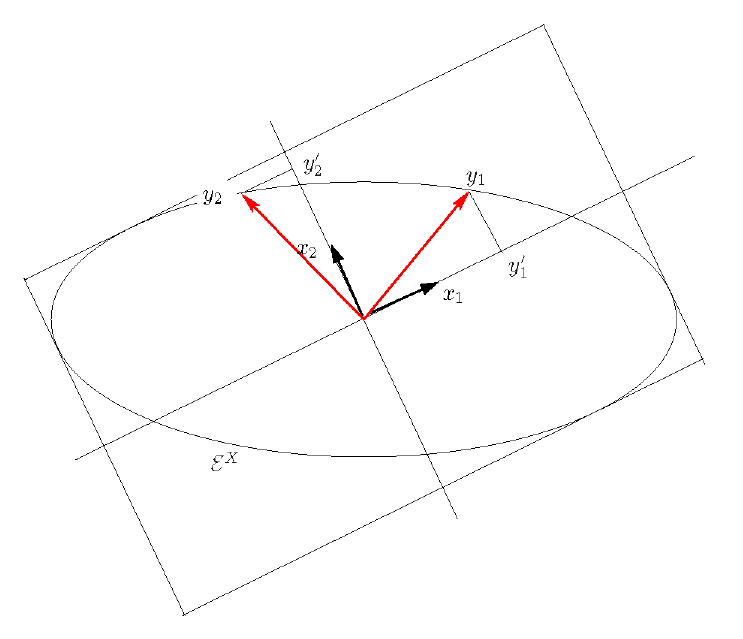}}}
\caption{Illustration of the upper bound to $\|Y\p\|_{F}^{2}$ in $\spann \{Y\}$ when $Y\p$ is defined by projection.}
\label{fig 2}
\end{center}
\end{figure}

Let $\EX$ be the ellipsoid of 
%the $\nl$-dimensional subspace 
$\spann(X)=\spann(Y)$ image by $A$ of the unit sphere of $\spann(Z)$.
%Let $\E = A\, \snvs$ be the $\ns$-dimensional ellipsoid  image by $A$ of the unit sphere $\snvs \subset \R^{\nvs}$, and~:
%\begin{equation}
%\label{800}\nonumber
%  \EX = \E \cap  \spann \{Y\} = \E \cap  \spann \{X\}  
%\end{equation}
%the $\nl$-dimensional ellipsoid, trace of $\E$ on the subspace spanned both by the given components $Y$ and the chosen basis $X$. 
By construction one has~:
\begin{equation}
\label{802}\nonumber
  y_{j} \in \EX \quad , \quad j=1 \dots \nl \ ,
\end{equation}
and the modified components $Y\p$ defined by projection satisfy, c.f. (\ref{532})~:
 \begin{equation}
\label{804}
   \|y\p_{j}\| = | \langle y_{j},x_{j}\rangle | \leq \ve_{j} \egaldef 
  \max_{\displaystyle y \in \EX} \langle y , x_{j}\rangle  \ , \  j=1\dots \nl \ ,
\end{equation}
so that~:
\begin{equation}
\label{805}
 \varproj^{X} Y \egaldef \|Y\p\|_{F}^{2} \leq \ve_{1}^{2}+ \dots+ \ve_{\nl}^{2} \ .
\end{equation}
We can now ``box'' the ellipsoid $\EX$ in the parallelotope $\PX$ of $\spann 
\{X\}$ defined by~:
\begin{equation}
\label{806}\nonumber
  \PX = \big \{  y\in \spann \{X\} \   | \  -\ve_{j} \leq \langle y,x_{j}\rangle \leq + \ve_{j}  \ , \  j=1\dots \nl \big \} \ ,
\end{equation}
(see figure \ref{fig 2}).
By construction, one can draw from each of the $2^{\nl}$ vertices of $\PX$ $\nl$ orthogonal hyperplanes tangent to the ellipsoid $\EX$, which implies that they are all on the orthoptic or Cartan sphere of the ellipsoid, whose radius is known to be the sum of the squares of the half principal axes $\sigma^{X}_{j}, j = 1 \dots \nl$ of $\EX$ (see for example the textbook \cite{PT2005}).

Hence~:
\begin{equation}
\label{808}
  \ve_{1}^{2}+ \dots + \ve_{\nl}^{2} = (\sigma^{X}_{1})^{2} + \dots +(\sigma^{X}_{\nl})^{2} \ .
\end{equation}
Let then $y^{X}_{1}\dots y^{X}_{\nl}$ be vectors whose extremity are points of $\EX$ located on its principal axes, so that~:
\begin{equation}
\label{810}\nonumber
 \|y^{X}_{j}\|=\sigma^{X}_{j}\ ,\ j=1\dots \nl \quad , \quad 
 \langle y^{X}_{i},y^{X}_{j}\rangle = 0\ , \ i,j=1\dots \nl , i\neq j \ .
\end{equation}
Property \eqref{546-1}  of Lemma \ref{lem 1} applied to the orthogonal components $Y=Y^{X}$ gives:~:
\begin{equation}
\label{812}
 (\sigma^{X}_{1})^{2} + \dots + (\sigma^{X}_{\nl})^{2} =  \|Y^{X}\|^{2} \leq
  \varsubspace \, Y^{X} \leq \varpca \ .
\end{equation}
Combining inequalities (\ref{805}) (\ref{808}) (\ref{812}) proves the inlem 4-2equality (\ref{534-2}).

%\subsection{Proof of Lemma \ref{lem 4-2}}  
%\label{proof of lem 4-2}
%
%
%
%When $ \varnorm^{X} Y$ attains its upper bound $\varpca$, \eqref{534-43} implies that $\|Y\p\|^{2}_{F} = \varsubspace Y\p$, which in turn implies by  \eqref{546-1} in Lemma \ref{lem 1} that $Z\p \egalperm \W_{\nl}$ and hence $X \egalperm \V_{\nl}$.

\subsection{Proof of Lemma \ref{lem 3-2} (notations of Section 
\ref{proof of lem 3}) }  
\label{proof of lem 3-2}

We prove first point 1 of the Lemma.
 Maximization of the convex function $X \leadsto \varproj Y^{X}$
 % = \sum_{j=1\dots\nl} \langle y_{j}\,,x_{j}\rangle^{2}$ 
 under the constraint $X^{T}X = I_{\nl}$  by algorithm \eqref{534-11}
 and passing to the limit proves \eqref{534-5}.

We prove now the ``only if'' part of point 2. Let $X$ be given such that 
$\varproj^{X} Y =\varpca$. Then necessarily~:
\begin{itemize}
  \item    equality holds in \eqref{812}, and property \eqref{546-2} of subspace variance
implies that the loadings $Z$ span the  subspace $\W_{\nl}$ of the $\nl$ first right singular vectors, which proves~\eqref{534-41}.
  \item equality holds in \eqref{804}, which implies that for $j\neq k$ the normals to $\EX$ at $y_{j}$ and $y_{k}$ are orthogonal (see Figure \ref{fig 2}). 
  The restriction $A_{\nl}$ of $A$ to $\spann \{\W_{\nl}\}$ is an isomorphism from $\spann \{\W_{\nl}\}$ to $\spann \{ \, \V_{\nl}\}$, hence~:
\begin{equation}
\label{814}\nonumber
\EX = \{ y \in \spann \{\, \V_{\nl}\} \  | \  \|A_{\nl}^{-1}y\|^{2} = 1 \} \ .
\end{equation}
 A normal $n(y)$ to $\EX$ at $y$ is then~:
\begin{equation}
\label{816}\nonumber
 n(y)= \half \nabla_{y} \big(  \|A_{\nl}^{-1}y\|^{2} - 1 \big)
 = (A_{\nl}^{-1})^{T}A_{\nl}^{-1}y =  (A_{\nl}^{-1})^{T} z \ ,
\end{equation}
and the orthogonality of $n(y_{j})$ and $n(y_{k})$ shows that~:
\begin{equation}
\label{819}
\langle n(y_{j}),n(y_{k})\rangle = \langle  (A_{\nl}^{-1})^{T} z_{1}, (A_{\nl}^{-1})^{T} z_{2}\rangle 
= \langle z_{1} , (A_{\nl}^{T}A_{\nl})^{-1} z_{2}\rangle=0\ ,
\end{equation}
which proves \eqref{534-42}. When vectors and matrices are written on the singular bases $\V_{\nl}$ and $\W_{\nl}$, one has 
$(A_{\nl}^{T}A_{\nl})^{-1} =  \diag\{ \frac{1}{\sigma_{1}^{2}} \dots \frac{1}{\sigma_{\nl}^{2}} \}$.
\end{itemize} 

We finally prove the ``if'' part of point 2. So let   \eqref{534-41} \eqref{534-42} \eqref{534-53} hold. Property   \eqref{534-41} implies that the half axes of $\EX$ are $\sigma_{1}\dots\sigma_{\nl}$, and \eqref{534-42}  that the normal $n_{j}$ to $\EX$ at $y_{j},j=1\dots\nl$ are orthogonal. 
So one can box $\EX$ with a parallelotope $\PX$ with axes parallel to the normals $n_{j}$, and define $X$ as the orthonormal basis along its axes. Then the same reasonning as above for the proof of (\ref{534-2}) shows that 
$\varproj Y = \varpca$, which ends the proof of the lemma.
%\cqfd

%\subsection{Proof of Lemma \ref{lem 4-2}}  
%\label{proof of lem 4-2}
%
%
%
%When $ \varnorm^{X} Y$ attains its upper bound $\varpca$, \eqref{534-43} implies that that $\|Y\p\|^{2}_{F} = \varsubspace Y\p $, which in turn implies by  \eqref{546-1} in Lemma \ref{lem 1} that $Z\p \egalperm \W_{\nl}$ and hence $X \egalperm \V_{\nl}$.
%%Possible rules for assigning a basis $X$ to the components $Y$ are~:
%

\subsection{Proof of Property \eqref{up proj pca formulation} }  
\label{proof of property up proj pca formulation}

Let the loadings $Z$ be such that 
%the components $Y=AZ$...
the UP-projected explained variance of the components $Y=AZ$ satisfies $\varprojup(Y)=\varpca$. Then by definition of $\varprojup$ one has~:
\begin{equation}
\label{ }
\varprojup(Y)=\varproj^{X}(Y) \text{ with } Y = UP, U^{T}U=I_{\nl}\ \ , \  \ 
P=P^{T}, P\geq 0 
\end{equation}
But $\varproj^{X}(Y)=\varpca$, and point 1 of Lemma \ref{lem 3-2} implies the existence of $P\p$ such that~:
\begin{equation}
\label{ }
%534-5
   Y \diag(X^{T}Y) = X P\p \text{ with ${P\p}^{T} \!= P\p$ and $P\p \geq 0$ }
\end{equation} 
Comparison of the two last properties shows that~:
\begin{equation}
\label{ }
Y = UP =  X P\p  \diag(X^{T}Y)^{-1} \ ,
\end{equation}
and uniqueness of the polar decomposition implies that
$P\p  \diag(X^{T}Y)^{-1} = P $ and hence is symmetrical, which can happen in only two cases~:
\begin{enumerate}
  \item either $\diag(X^{T}Y) = \lambda I_{\nl}$ for some $\lambda$, which gives the parasitic solution $Z = Z^{\#}$,
  \item or $P\p$ - and hence also $P$ itself - is diagonal, which implies that the components $Y=UP$ are orthogonal. But point 2 of Lemma \ref{lem 3-2} implies that they are also $(A^{T}_{\nl}A_{\nl})^{-1}$-- orthogonal, which is possible only if the components $Y$ are proportional to the left singular basis $\V_{\nl}$, which gives the SVD solution $Z \egalperm \W_{\nl}$. %\cqfd
\end{enumerate}

\bibliographystyle{plainnat}
\bibliography{Companion_paper_one}% common bib file

\end{document}